\documentclass{article}

 \usepackage[dblblindworkshop, final]{neurips_2025}
\usepackage[utf8]{inputenc} % allow utf-8 input
\usepackage[T1]{fontenc}    % use 8-bit T1 fonts
\usepackage{hyperref}       % hyperlinks
\usepackage{url}            % simple URL typesetting
\usepackage{booktabs}       % professional-quality tables
\usepackage{amsfonts}       % blackboard math symbols
\usepackage{nicefrac}       % compact symbols for 1/2, etc.
\usepackage{microtype}      % microtypography
\usepackage{xcolor}  % colors
\usepackage{algorithm}
\usepackage{algpseudocode}
\usepackage{graphicx}
\graphicspath{{figures/}}
\usepackage{amsmath}
\usepackage{natbib}
\workshoptitle{Efficient Reasoning}

\title{{\texttt{ProofSketch}}: Efficient Verified Reasoning for Large Language Models}

\author{
Disha Sheshanarayana\thanks{These authors contributed equally to this work} \\
Manipal University Jaipur\\
\texttt{disha.229301161@muj.manipal.edu}
\And
Tanishka Magar\footnotemark[1] \\
Manipal University Jaipur\\
\texttt{tanishka.229301736@muj.manipal.edu}
}
\begin{document}

\maketitle

\begin{abstract}
 Reasoning methods such as chain-of-thought prompting and self-consistency have shown immense potential to improve the accuracy of large language models across various reasoning tasks. However such methods involve generation of lengthy reasoning chains, which substantially increases token consumption, computational cost, and latency. To address this inefficiency, we propose \texttt{ProofSketch}, a verification-guided reasoning framework that integrates symbolic closure computation, lexicographic verification and adaptive sketch generation. Our experiments show that \texttt{ProofSketch} consistently reduces token usage while improving accuracy, demonstrating that this approach offers a promising path for efficient and trustworthy reasoning. The code is available at https://github.com/tanishka66/ProofSketch.
\end{abstract}

\section{Introduction}
The reasoning capabilities of LLMs have been explored, analysed and experimented with. Research on LLM reasoning spans prompting-based strategies, structured search methods, and efficiency-oriented decoding, each offering different trade-offs between accuracy, interpretability, and cost. However, these approaches typically require the model to generate longer reasoning chains. While this can improve accuracy, it also comes at the cost of substantial token usage and higher latency, reducing their efficiency in settings where computational resources are limited.
\par
It has been seen that in many cases, LLMs generate unnecessarily long and elaborate reasoning chains, even for trivial questions, often referred to as the overthinking problem~\cite{yue2025dont}. This leads to significant wastage of computational resources.
Previous works have attempted to improve upon this by enforcing constraints, using instruction tuning and dynamically adjusting reasoning length based on problem difficulty to control output lengths~\cite{han2025token,zhao2025saber}.
Another limitation is that intermediate reasoning steps remain unchecked. ~\cite{nguyen2024direct} Long reasoning traces provide no guarantee of logical validity. As a result, we cannot be certain whether the final prediction is grounded in valid reasoning or reflect the errors propagated through fluent but incorrect intermediate steps.
Combined, these challenges often cause excessive reasoning length and lack of intermediate verification. This highlights the need for methods that can achieve accurate reasoning under tight compute budgets while also providing correctness guarantee. 
\par
To address this gap we propose \texttt{ProofSketch}, a verification-guided efficient reasoning framework. Instead of generating lengthy reasoning chains, our method produces multiple short "sketches" containing atomic claims, then selects the sketch with maximum verification coverage, thus enabling low-cost yet reliable shallow reasoning.

\section{Related Works}
Building on the growing interest in improving LLM reasoning, scientists have proposed techniques such as chain-of-thought (CoT) prompting, self-consistency, and proof generation, which have achieved notable increases in accuracy~\cite{wei2022chain,wang2023selfconsistency,yang2022generating}, but computational costs remain excessive. To overcome this studies have explored reasoning under explicit budget constraints applying token-aware optimization strategies~\cite{han2025token}, adaptive control of reasoning length via confidence probing~\cite{fu2025reasoning}, and dynamic token budgeting that adjusts the reasoning process per instance~\cite{li2025selfbudgeter}. Other methods to reduce output lengths include replacing token-heavy CoT traces with soft latent vectors~\cite{xu2025softcot} or leveraging reward shaping to encourage more concise reasoning.
\par
The study~\cite{zhang2025atomic} showed how reducing tasks to minimal (atomic) claims can improve soundness and interpretability, which directly inspired \texttt{ProofSketch}. Works ~\cite{aytes2025sketch} and~\cite{wang2025speculative} have explored sketching-based methods where shorter reasoning structures are generated, successfully reducing token usage. Studies have also explored ways to ensure correctness in responses. One of them which stands out is verification-guided reasoning, where intermediate steps are explicitly checked~\cite{yang2022generating,ling2023deductive,cao2024graphreason,chowdhury2025zero}. ProofSketch combines these strands to ensure verified, token-efficient reasoning.

\section{\texttt{ProofSketch}}

\subsection{Problem Formulation}

Let $\mathcal{T}$ be a logical theory containing facts and rules, and $Q$ be a question requiring True/False/Unknown classification. Given the computational constraints of modern language models, we seek reasoning methods that simultaneously optimize multiple objectives: generating accurate answers, minimizing computational overhead, and providing formal guarantees about reasoning correctness.

We define the efficiency-accuracy-certification optimization problem as: $\max$ $\text{Accuracy}(f(\mathcal{T},Q))$ $\text{s.t.}$ $\mathbb{E}[\text{tokens}(f(\mathcal{T},Q))] \leq \beta$, $\text{Cert}(f(\mathcal{T},Q)) \to 1$ where $f$ is our reasoning function, $\beta$ is the token budget constraint, and $\text{Cert}$ measures the proportion of formally verified reasoning steps. This multi-objective formulation captures the fundamental challenge in neural reasoning: balancing accuracy, efficiency, and reliability while providing formal guarantees about reasoning processes.

\subsection{{\texttt{ProofSketch}} Framework}

\begin{figure}[t]
    \centering
    \includegraphics[width=\linewidth]{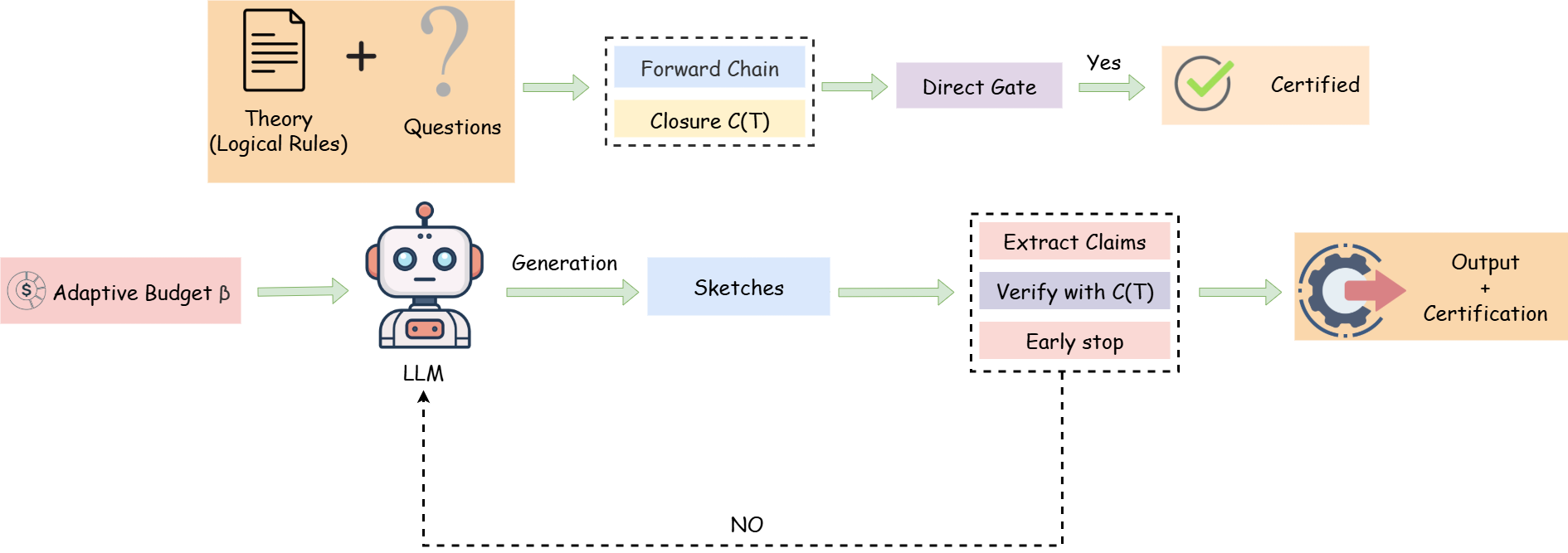}
    \caption{\texttt{ProofSketch} framework generates sketches from theories and questions, applies closure verification through forward chaining, and uses adaptive budgeting for re-generation when verification fails. Successful verification produces certified outputs, while failures trigger budget-controlled iterative refinement until certification is achieved.}
    \label{fig:proofsketch_flow}
\end{figure}

\texttt{ProofSketch} introduces a novel reasoning framework that integrates symbolic closure computation with verifier-gated neural generation. The framework operates through a multi-stage pipeline designed to optimize accuracy, efficiency, and certification simultaneously, as illustrated in Figure~\ref{fig:proofsketch_flow}.

\paragraph{Symbolic Closure Foundation}
The framework begins by parsing theory $\mathcal{T}$ into positive facts $F_+$, negative facts $F_-$, and logical rules $R$. Forward chaining derives the symbolic closure $C(\mathcal{T}) = \text{FC}(F_+, F_-, R)$, which serves as the foundation for both direct answer checking and claim verification. This closure enables immediate resolution of questions derivable through pure logic while providing a verification oracle for generated content.

\paragraph{Verifier-Gated Generation}
When generation is required, the framework samples up to $K{=}4$ short \emph{sketch} candidates under an adaptive token budget (120 tokens if $C(\mathcal{T})$ already contains facts about the queried entity, 160 otherwise) with mild temperature ($\tau{=}0.3$) for controlled diversity. Each sketch is emitted in a structured, machine-readable format consisting of a proposed answer to the query $Q$ together with a small set of atomic declarative claims; each claim must conform to the canonical unary form ``$e$ is $a$'' or ``$e$ is not $a$'', where $e$ is an entity and $a$ is an attribute. We canonicalize surface mentions so that entities and attributes align with the symbols in $\mathcal{T}$ and reduce the sketch to a minimal anchored subset that refers directly to the entity named in $Q$, yielding a compact, query-focused justification rather than an arbitrary chain-of-thought. Because model outputs may be imperfectly structured, we apply a lightweight repair pass prior to parsing; if a sampled sketch cannot be reliably parsed into a candidate answer plus at least one canonical claim, it is treated as having no usable claims and is de-prioritized by the verifier-guided selector (i.e., it cannot be marked certified downstream).

\paragraph{Multi-Objective Verification and Selection}
Generated sketches are evaluated through formal verification against $C(\mathcal{T})$. The framework employs lexicographic scoring that prioritizes: (1) full certification (all claims verified), (2) partial verification coverage, (3) token efficiency, and (4) consistency with closure decisions. Early stopping occurs when a fully certified sketch is found, providing computational savings while ensuring formal correctness guarantees.

\paragraph{Certification and Output}
The framework produces three key outputs: a final answer (with closure correction if needed), a set of formally verified atomic claims supporting the reasoning, and a certification status indicating the degree of formal verification achieved. This design enables transparent reasoning assessment and post-hoc analysis of model decisions.

\section{Experimentation}
\subsection{Experimental Setup}

\textbf{Datasets.} We evaluate \texttt{ProofSketch} on a curated subset of the ProofWriter dataset~\cite{tafjord2020proofwriter}, a widely-used logical reasoning benchmark containing theories expressed in natural language with corresponding True/False/Unknown questions. We collected 300 data points from ProofWriter, selected to represent varying reasoning depths and complexity levels to ensure comprehensive evaluation of our symbolic-neural integration approach.

\textbf{Baselines.} We evaluate three prompting approaches across three language models: Mistral-7B~\cite{jiang2023mistral}, DeepSeek-R1 Distill~\cite{deepseek2025r1} Llama-8B, and Qwen-7B. The prompting approaches include: (1) \textbf{Zero-shot} prompting for immediate True/False/Unknown classification without reasoning steps, (2) \textbf{Short CoT} with up to 3 concise reasoning lines before the final answer, and (3) \textbf{Long CoT} allowing up to 10 detailed reasoning steps.

\textbf{Metrics.} We measure accuracy on True/False/Unknown classifications, certification rate (percentage of formally verified responses), mean token usage, P95 token consumption, and average latency. 

\textbf{Implementation Details.} \texttt{ProofSketch} uses $K=4$ sketch candidates with adaptive token budgets ($\beta_1=120$ for entities in symbolic closure, $\beta_2=160$ otherwise) and temperature $\tau=0.3$ for controlled diversity. The forward-chaining engine computes symbolic closures using first-order logic predicates. Multi-objective scoring employs lexicographic ordering across certification status, verification coverage, token efficiency, and response consistency.

\subsection{Results}

We evaluate \texttt{ProofSketch} across three language models on a 300 example ProofWriter dataset, measuring accuracy, token efficiency, and formal verification capabilities. Table~\ref{tab:results-main} presents our comprehensive evaluation results, demonstrating that \texttt{ProofSketch} achieves strong reasoning performance across all models: 68.0\% accuracy with R1-Distill-Llama-8B, 52.0\% with Mistral-7B, and 54.0\% with R1-Distill-Qwen-7B. Beyond competitive accuracy, a key distinguishing feature of our method is its formal verification capability, entirely absent in standard prompting approaches. The method achieves remarkable certification rates of 42.0\% with R1-Llama and R1-Qwen, and 84.0\% with Mistral-7B, representing responses that receive complete mathematical verification through our symbolic closure system. Regarding computational efficiency, \texttt{ProofSketch} demonstrates exceptional performance with Mistral-7B, requiring only 27.96 tokens per query on average, R1-Qwen achieving similar efficiency at 30.28 tokens, while the R1-Llama configuration uses 137.94 tokens but delivers the highest accuracy. These results validate our core hypothesis that symbolic preprocessing and verifier-gated generation enable both competitive reasoning performance and formal verification capabilities, establishing a new paradigm for trustworthy neural reasoning systems while maintaining computational efficiency across different model architectures. We have analyzed the detailed findings in Appendix~\ref{app:stat}. To assess the generalizability of our findings, we additionally conduct evaluations on a larger 1,000 example ProofWriter dataset, with full results reported in Appendix~\ref{app:extended-eval}.

\begin{table}[H]
  \caption{Comparison of reasoning methods on three models (Deepseek-R1-Distill-Llama-8B, Mistral-7B, Deepseek-R1-Distill-Qwen-7B), reporting Accuracy(Acc), Mean tokens(Tok), and Certified fraction(Cert).}
  \label{tab:results-main}
  \centering
  \begin{tabular}{l *{3}{ccc}}
    \toprule
    & \multicolumn{3}{c}{R1-Distill-Llama-8B} 
    & \multicolumn{3}{c}{Mistral-7B} 
    & \multicolumn{3}{c}{R1-Distill-Qwen-7B} \\
    \cmidrule(lr){2-4} \cmidrule(lr){5-7} \cmidrule(lr){8-10}
    Method & Acc & Tok & Cert & Acc & Tok & Cert & Acc & Tok & Cert \\
    \midrule
    Zero-shot    & 0.37 & \textbf{122.27} & 0    & 0.33 & \textbf{7.00}  & 0    & 0.39 & \textbf{9.85}  & 0 \\
    Short-CoT    & 0.52 & 214.83 & 0    & 0.48 & 52.86 & 0    & 0.44 & 48.75 & 0 \\
    Long-CoT     & 0.52 & 218.71 & 0    & 0.41 & 101.76& 0    & 0.47 & 101.09& 0 \\
    \texttt{ProofSketch}  & \textbf{0.68} & 137.94 & \textbf{0.42} & 0.52 & 27.96 & \textbf{0.84} & \textbf{0.54} & 30.28 & \textbf{0.42} \\
    \bottomrule
  \end{tabular}
\end{table}

\begin{figure}[H]
    \centering
    \includegraphics[width=\linewidth]{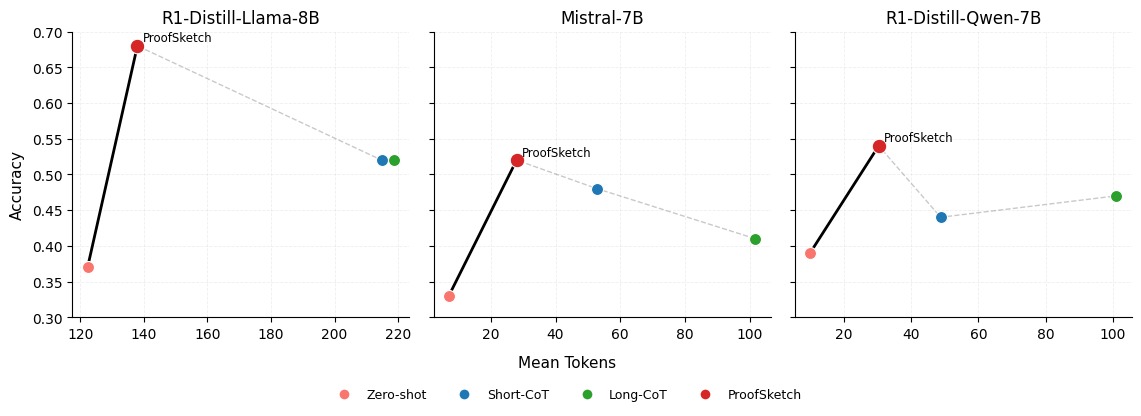}
    \caption{Pareto Frontier for Accuracy vs Token Usage across Models}
    \label{fig:pareto}
\end{figure}

\subsubsection{Comparison of Latency}

Latency results across all methods and models are provided in  Table~\ref{tab:results-latencies}. While ProofSketch consistently reduced token usage compared to other prompting techniques, our results reveal that this efficiency sometimes could lead to a noticeable latency overhead. Our framework remained competitive for most and even significantly reduced on Mistral-7B. The only notable increase in the latency overhead was for R1-Distill-Llama-8B. We did not identify a single dominant cause for this increase, but we hypothesize that the cumulative costs of multiple sketch generations, closure computation, and verification checks outweigh the generation savings in some cases. Future work could explore caching closure computations and parallelizing sketch generation to mitigate and control this overhead.

\begin{table}[H]
  \caption{Mean latency (ms) across reasoning strategies and models.}
  \label{tab:results-latencies}
  \centering
  \begin{tabular}{lccc}
    \toprule
     Method & Distill-Llama-8B & Mistral-7B & Distill-Qwen-7B \\
    \midrule
    Zero-shot    & 9240.05  & 1018.62  & 1340.26 \\
    Short-CoT    & 15908.20 & 7581.99  & 6926.19 \\
    Long-CoT     & 15984.95 & 11069.88 & 9909.83 \\
    \texttt{ProofSketch}  & 31741.47 & 5593.77  & 11153.46 \\
    \bottomrule
  \end{tabular}
\end{table}

\section{Conclusion}
\textbf{(1) Conclusion.} In this work we propose \texttt{ProofSketch}, a novel framework designed to ensure efficient and verified reasoning of large language models by combining symbolic closure, adaptive sketch generation, and lexicographic verification. ProofSketch involves the generation of multiple short sketches with atomic claims, which are then verified to select the most reliable sketch. This directly addresses two key limitations of prior approaches: longer reasoning lengths and unverified intermediate steps and thus offers a promising direction for deploying LLMs in compute-constrained or tightly budgeted settings.
\textbf{(2) Limitations.} The additional verification stage in \texttt{ProofSketch} introduces a modest latency overhead compared to purely generative baselines. A key limitation of \texttt{ProofSketch} is that it relies on simple symbolic checks, which may not scale to more complex reasoning domains. Furthermore, it has only been tested on controlled datasets, meaning its effectiveness in real-world noisy environments remains to be determined. 
\textbf{(3) Future Work.} Future work could involve extending this framework to more complex domains, exploring adaptive sketch generation policies, and integrating neural verifiers for broader coverage.

\bibliographystyle{plain}
\bibliography{references}

\appendix
\setlength{\textfloatsep}{5pt}   % space below floats
\setlength{\floatsep}{5pt}       % space between floats
\setlength{\intextsep}{5pt}      % space above/below in-text floats

\section{{\texttt{ProofSketch}} Algorithm}

\begin{algorithm}[H]
\caption{{\texttt{ProofSketch}} ProofSketch with Verifier-Gated Decoding}
\label{alg:proofsketch}
\textbf{Input:} Theory $\mathcal{T}$, question $Q$, sketches $K=4$, budgets $\beta_1=120, \beta_2=160$\\
\textbf{Output:} Answer $\hat{a}$, verified claims $\hat{C}$, certification status $\sigma$
\begin{algorithmic}[1]
\State Parse $\mathcal{T}$ into facts $(F_+, F_-)$ and rules $R$
\State $C(\mathcal{T}) \leftarrow \text{ForwardChain}(F_+, F_-, R)$
\If{$Q \in C(\mathcal{T})$}
    \State \textbf{return} $(\text{closure\_answer}(Q), \emptyset, \text{CERTIFIED})$
\EndIf
\State $\beta \leftarrow \beta_1$ if $\text{entity}(Q) \in C(\mathcal{T})$ else $\beta_2$
\For{$k = 1, 2, \ldots, K$}
    \State $s_k \leftarrow \text{LLM}(\text{prompt}(\mathcal{T},Q), \beta, \tau)$
    \State $(\text{claims}_k, \text{answer}_k) \leftarrow \text{parse\_json}(s_k)$
    \State $\text{verdicts}_k \leftarrow [\text{verify}(c, C(\mathcal{T})) \mid c \in \text{claims}_k]$
    \State $\text{score}_k \leftarrow (\text{cert}_k, |\text{verified}_k|, -\text{tokens}(s_k), \text{consistency}_k)$
    \If{$\text{cert}_k = 1$}
        \State \textbf{return} $(\text{answer}_k, \text{claims}_k, \text{CERTIFIED})$
    \EndIf
\EndFor
\State $k^* \leftarrow \arg\max_{\text{lex}}(\text{score}_k)$
\If{$C(\mathcal{T}) \models Q$}
    \State $\hat{a} \leftarrow \text{closure\_answer}(Q)$
\Else
    \State $\hat{a} \leftarrow \text{answer}_{k^*}$
\EndIf
\State \textbf{return} $(\hat{a}, \text{verified\_claims}_{k^*}, \text{certification\_status})$
\end{algorithmic}
\end{algorithm}

\section{Statistical Analysis}

\label{app:stat}
\begin{figure}[H]
    \centering
    \begin{minipage}{0.48\linewidth}
        \centering
        \includegraphics[width=\linewidth]{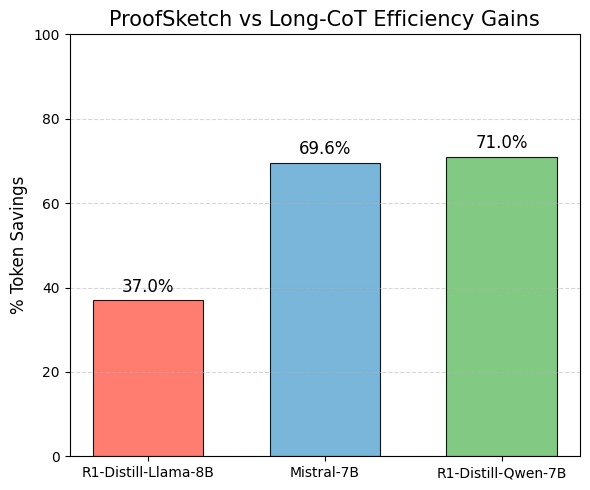}
        \caption{Token Savings}
        \label{fig:eff}
    \end{minipage}\hfill
    \begin{minipage}{0.48\linewidth}
        \centering
        \includegraphics[width=\linewidth]{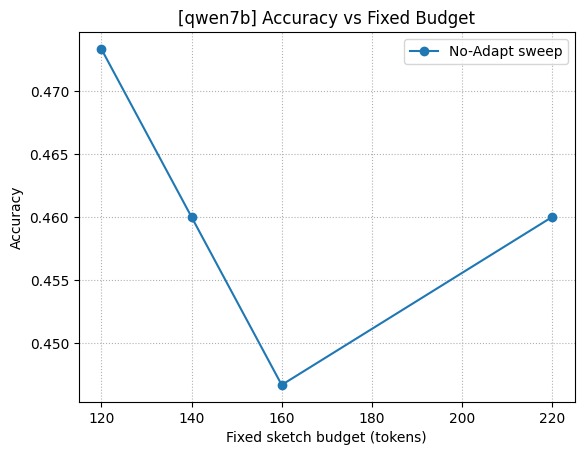}
        \caption{Ablation Study}
        \label{fig:ablation}
    \end{minipage}
\end{figure}

To demonstrate \texttt{ProofSketch}'s efficiency advantages, we analyzed token savings compared to Long-CoT baselines across all three models. Figure~\ref{fig:eff} shows that ProofSketch achieves substantial computational savings: 37.0\% token reduction on R1-Distill-Llama-8B, 69.6\% on Mistral-7B, and an impressive 71.0\% on R1-Distill-Qwen-7B. Notably, the Qwen-7B model demonstrates the highest efficiency gains, indicating optimal compatibility with our symbolic verification approach. 

To validate our adaptive budgeting mechanism, we conducted an ablation study on R1-Distill-Qwen-7B examining the impact of fixed sketch budgets versus our adaptive allocation strategy. Figure~\ref{fig:ablation} reveals that fixed budget constraints lead to suboptimal performance across the entire budget range (120-220 tokens), with accuracy consistently remaining below the adaptive approach. This demonstrates that our dynamic budget allocation is crucial for optimal performance, as it intelligently adjusts computational resources based on problem complexity rather than applying rigid constraints. These findings confirm that \texttt{ProofSketch}'s integrated approach delivers superior efficiency while maintaining reasoning quality through intelligent resource management.

\section{Extended Evaluation}
\label{app:extended-eval}
To assess the generalizability of our findings, we further evaluate \texttt{ProofSketch} on an extended 1,000 example subset ProofWriter dataset. We saw that the overall performance trends remained consistent, thus confirming that the \texttt{ProofSketch} framework can scale on a larger dataset effectively without degradation in accuracy and tokens. These results suggest that the observed benefits of \texttt{ProofSketch} are not limited to small-scale benchmarks but extend to broader settings as well.

\begin{table}[H]
  \caption{Evaluation of \texttt{ProofSketch} across larger dataset}
  \label{tab:results-latency}
  \centering
  \begin{tabular}{lccc}
    \toprule
     Method & Accuracy & Mean Tokens & Mean Latency \\
    \midrule
    Zero-shot    & 0.394  & 9.598  & \textbf{1312.06} \\
    Short-CoT    & 0.404 & 49.735  & 7042.52 \\
    Long-CoT     & 0.424 & 98.126 & 10012.91 \\
    \texttt{ProofSketch}  & \textbf{0.496} & \textbf{28.622}  & 9468.45 \\
    \bottomrule
  \end{tabular}
\end{table}

\end{document}